# AUTOMATED FINGERPRINT RECOGNITION: USING MINUTIAE MATCHING TECHNIQUE FOR THE LARGE FINGERPRINT DATABASE

*S M Mohsen, S M Zamshed Farhan and M M A Hashem*

Department of Computer Science and Engineering
Khulna University of Engineering & Technology
Khulna-9203, Bangladesh
Ph: +88-041-769471, Fax: +88-041-774403
Email: hashem@cse.kuet.ac.bd

## ABSTRACT

Extracting minutiae from fingerprint images is one of the most important steps in automatic fingerprint identification system. Because minutiae matching are certainly the most well-known and widely used method for fingerprint matching, minutiae are local discontinuities in the fingerprint pattern. In this paper a fingerprint matching algorithm is proposed using some specific feature of the minutiae points, also the acquired fingerprint image is considered by minimizing its size by generating a corresponding fingerprint template for a large fingerprint database. The results achieved are compared with those obtained through some other methods also shows some improvement in the minutiae detection process in terms of memory and time required.

## 1. INTRODUCTION

Forensic experts [1] usually look for a sufficient number of matching *minutiae* in order to determine whether two fingerprint images correspond to the same person. Automatic matching algorithms roughly follow the same procedure, according to the dominant approaches reported in the literature. In order to promote the wide spread utilization of biometric techniques, an increased level of security of biometric data is necessary [3]. Minutiae characteristics are local discontinuities in the fingerprint pattern which represent terminations and bifurcations [Fig. 1]. [2]. There are a lot of hindrances make very hard the fingerprint verification task and very robust algorithms need to process the noisy fingerprint images. An automatic verification system is proposed here. The system has to recognize an acquired fingerprint image using personal data information to select a fingerprint database item. A fingerprint matching is successively performed between the acquired fingerprint image and the selected database item. A lot of works dealing with fingerprint verification task is present in literature [5]. They obtained high accuracy of correct verification but they did not work with poor and the large size of image. In this paper a minutiae matching algorithm is proposed for fingerprints verification, which is based on a set image processing algorithms and characteristic features extraction algorithms [4]. The proposed system is based on three main phases: the image processing phase, the minutiae extraction phase and the matching phase. An image processing technique is used to erase noisy background, textures of the fingerprint. The minutiae extraction phase aims to extract fingerprint images characteristics. Finally, the matching phase determines whether the fingerprints are impressions of the same finger. The matching stage also defines a threshold to decide whether given pair of representations is of the same finger (mated pair) or not. The rest of the paper is organized as follows: Section 2 addresses the details of the proposed system that is image enhancement, minutiae extraction technique, and matching stage. Section 3 presents the experimental results and performance evaluation of our proposed approach. Finally, in Section 4 some conclusions are drawn.

## 2. THE PROPOSED SYSTEM

The goal of a this system is to compare two fingerprint images, where a fingerprint is composed of two main features which are called minutiae points can be classified as ridge ending and bifurcation.

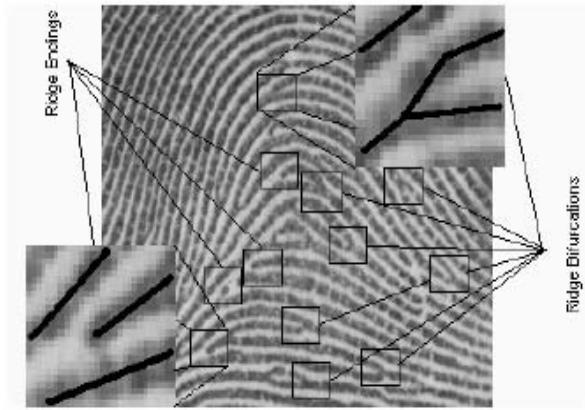

**Fig. 1**: Ridge ending and bifurcation

A fingerprint captured and processed in a particular moment with one fingerprint stored in the database in real time. In this work an automated fingerprint verification system is proposed. For matching there are two steps, a registration and verification step. In the registration step all fingerprints of a set of person are captured, processed and stored as a template which is much smaller than the original image, in a database for later use. In the verification step a person gives his fingerprint to verify the identity: the fingerprint is compared with ones stored in the database

### 2.1 Image Processing

Image processing consists of three stages: 1. Image Filtering, 2. Image Enhancement, and 3. Image Shaping. In Image Filtering every image is a collection of pixels, where each pixel contains three colors in different ratio, range over 0 to 255 RGB values, which has three parameters, such as RED, GREEN and BLUE as constant figure. For an example, return value as RGB of all 0 contains BLACK and return value as RGB of all 255 contains WHITE. For filtering RGB values of every pixel is replaced by maximum or minimum, with comparing the given threshold. In Image Enhancement, the black colors mean the ridgelines, which haven't the same thickness all over the image. To overcome the above limitations find the representative point of each ridgeline and fill it. In Image Shaping, it requires lining the image which is the initial step to overcome the limitation of smoothness of the image. Then from lined image, each single BLACK pixel is found out and draws a filled circle corresponding to the pixel to smooth the enhanced image.

### 2.2 Minutiae extraction and Template Generation

Minutiae extraction is just a trivial task of extracting singular points in a thinned ridge map. The performance of currently available minutiae extraction algorithms depends heavily on the quality of input fingerprint images. After extracting the minutiae points a predetermined position is clipped to generate a template which bears some sophisticated data to generate a corresponding data file used as template, which file contains the Total number of minutiae points, Co-ordinates of each point. Interrelated distance of each point termed as correlation.

### 2.3 Fingerprint Matching

Matching is the final stage of this study. It is needed to verify one that he must be registered before. So the steps are: **1**.Registration, **2**.Verification.

*Registration:* In this process, the process takes the one's fingerprint as an image format and processed that image as few steps such as filtering, enhancing, lining and shaping. Then it requires selection of minutiae points as feature then generates a template and stores it. The authenticate template contains the total number of minutiae points selected by proposed feature selection technique AFIS. The number of minutiae points limited by a bindings named limited region to improve the flexibility of verification, and then find out the co-relation among the features bounded by limited region.

*Verification:* To verify one, the process takes one's fingerprint as an image format through fingerprint acquisition hardware and processed that. Then it requires technique to detect minutiae points and selects features, and then to verify, load the templates and compare with the information gathered from verifying one. If it obtains any template matched with that of verifying one, it makes a decision that one was authenticated, or not.

### 2.4 Algorithm: Matching

1. *Take X and Y as two co-ordinates.*
2. *Take a minutiae point (I th), where 1<=I<=total minutiae points.*
3. *Now, take another minutiae point J th,*
   *Where 1<=J<=total minutiae points*
   *and I!=J.*
4. *if Y_I=Y_J then DISTANCE: =X_J-X_I*
   *if DISTANCE <0 then*
   *DISTANCE:=DISTANCE×(-1);*
   *else if (Y_I>Y_J) then*
   *if (Y_I-Y_J:=1) or (Y_I-Y_J:=-1)*

```
    then  DISTANCE:=0
else if (Y_I-Y_J:=2) or (Y_I-Y_J:=-2) then
        DISTANCE:=Y_I-Y_J
else    DISTANCE:=(Y_I-Y_J)-2
        DISTANCE:=(DISTANCE*WIDTH)
            +(WIDTH-X_J)+X_I
 else if (Y_I-Y_J):=1 or(Y_I-Y_J):=-1
   then  DISTANCE:=0
else if (Y_I-Y_J:=2) or (Y_I-Y_J:=-2)
   then DISTANCE:=Y_J-Y_I
else DISTANCE:=(Y_J-Y_I)-2
DISTANCE := (DISTANCE*WIDTH)
            +(WIDTH-X_I)+X_J
```
5. Repeat step 1,2 & 3 for the verified image also.
6. Find the equality of minutiae points (M1 & M2), as EQ: = matched (M1, M2)/greater (M1, M2);
7. Calculate total number of correlated distances DISTANCE-of-both.

## 3. EXPERIMENTAL RESULTS

**Table 1:** Classifications of Minutiae point of an image.

| | Image | 1 | 2 | 3 | 4 |
|---|---|---|---|---|---|
| Minutiae | Contained in image | 54 | 37 | 32 | 30 |
| | Selected by AFIS | 44 | 43 | 38 | 32 |
| | Dropped by AFIS | 22 | 12 | 12 | 9 |
| | False by AFIS | 12 | 18 | 18 | 11 |
| | Correct selection | 32 | 25 | 20 | 21 |

In table 1, it contains some particular images with manually detected minutiae points contained in the image, Minutiae selected by the proposed technique AFIS, Dropped minutiae points by AFIS, False minutiae points by AFIS, and Correct minutiae points acquired by AFIS. Total minutiae points contained in taken image calculated manually are used to find out the accuracy of proposed minutiae finding technique named AFIS. AFIS is a proposed minutiae finding technique used to find out the fingerprint feature named minutiae points from processed image artificially.

Dropped minutiae points are that which are contained in taken image but not in the image processed by the AFIS, i.e., failed to recognize that point as minutiae points or as feature. False minutiae points are that which are not contained in taken image as minutiae points but the AFIS select that as minutiae points. Then correct numbers of minutiae points are selected by AFIS are those features, which are contained in original fingerprint image detected by AFIS.

**Table 2:** Percentage of different stages of minutiae points

| | Image | | 1 | 2 | 3 | 4 |
|---|---|---|---|---|---|---|
| Percentage of | Total Minutiae | | 54 | 37 | 32 | 30 |
| | False | Minutiae by AFIS | 22.22 % | 48.69 % | 56.22 % | 36.67 % |
| | Drop | | 40.74 % | 32.43 % | 37.5 % | 30.00 % |
| | Correct | | 59.26 % | 67.57 % | 62.5 % | 70.00 % |

In table 2, it contains the total number of minutiae points contained in taken image, percentage of false minutiae points by AFIS, percentage of dropped minutiae points by AFIS and percentage of correct minutiae points by AFIS.

### 3.1    Performance Evaluation

The poor image means the original image and then extracted minutiae points by AFIS from original poor image. The improved image is the processed image, extracted minutiae points by AFIS from improved image

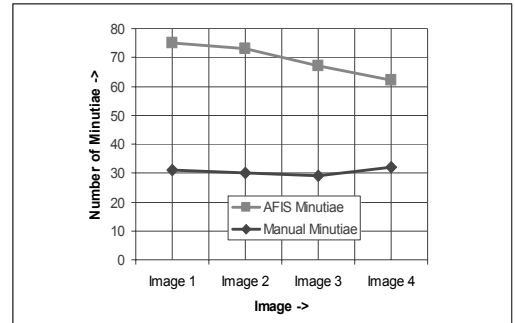

**Fig. 2:** Graph represents the poor versus improved image curve

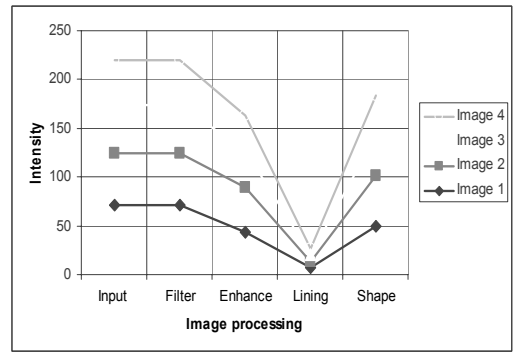

**Fig. 3:** Graph represents the ridgeline intensity curve during the various phases of the input image processing

In Fig. 2, it is shown that the number of minutiae points of poor image and improved image is not similar but linear and it takes a decision that number of minutiae points in poor image is less than that of the improved image, as the poor image is so noisy compared with that of the improved images.

In Fig. 3, it shows that the intensity variation and relationship among the intensity variation of the steps, such as input image, filtered image, enhanced image, lined image and shaped image.

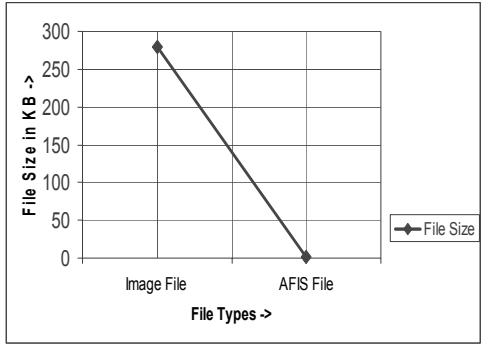

**Fig. 4:** Graph represents image file size Vs AFIS generated file size

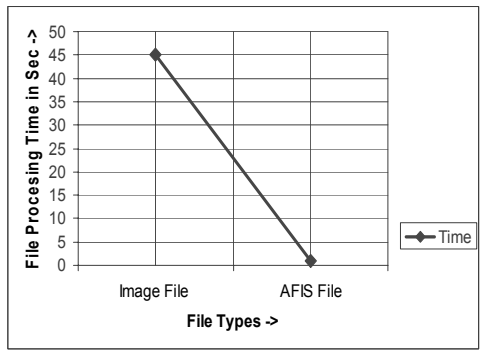

**Fig. 5:** Graph represents processing time of image file Vs AFIS generated file size

In Fig. 4, it represents the memory management criterion of AFIS, where processed image requires a large amount of storage area, such as 280 kilobytes, on the contrary, the AFIS generated file requires too less memory, such as 1 kilobyte.

In Fig. 5, it represents the computational time of AFIS, where processed image requires a large amount of computational time, such as 45 second, on the contrary, the AFIS generated file requires too less computational time, such as 1 second.

## 4. CONCLUSION

Every person or organization search for more secure authentication methods to secure one. A Fingerprint based security system can provide more secure and reliable authentication and verification according users requirements. In this paper a matching algorithm is provided to authenticate and verify one. The proposed approach comparatively shows more efficiency on the case of memory and time requirements. This matching approach can be improved by taking whole image in place of representative part of image.

## REFERENCES

[1] FBI. The science of fingerprints. Classification and uses. U.S. Department of Justice, FBI. Superintendent of Documents, U.S. Government Printing Office, Washington, D.C. 20402. Stock Number 027-001-00033-5 / Catalogue No. J1.14/2:F49/12/977, 1984.

[2] K. Jain, L. Hong, S. Pankanti, R. Bolle, An identity-Authentication system using fingerprints, Proceeding of IEEE, Vol.85, No.9, 1997, pp.1365-1388.

[3] N.K. Ratha, J.H. Connell, and R.M. Bolle, "An analysis of minutiae matching strength", *Proc. Third AVBPA*, Halmstad, Sweden, June 2001, pp. 223-228.

[4] S M Mohsen, S M Zamshed Farhan, M M A Hashem, K M Azharul Hasan and Banani Roy, Fingerprint Recognition Using a Feature Selection Technique, proc. 3$^{rd}$ International Conferences on Electrical, Electronics and Computer Engineering (ICEECE 2003), pp 171-176.

[5] Z. M. Kovacs-Vajna, A fingerprint verification system based on triangular matching and dynamic time warping, IEEE transaction on pattern analysis and machine intelligence, Vol.22, No.11, November 2000, pp.1266-1276.